# Restricting Greed in Training of Generative Adversarial Network

Haoxuan You, Zhicheng Jiao, Haojun Xu, Jie Li, Ying Wang, Xinbo Gao

Xidian University

## Abstract

*Generative adversarial network (GAN) has gotten wide re-search interest in the field of deep learning. Variations of GAN have achieved competitive results on specific tasks. However, the stability of training and diversity of generated instances are still worth studying further. Training of GAN can be thought of as a greedy procedure, in which the generative net tries to make the locally optimal choice (minimizing loss function of discriminator) in each iteration. Unfortunately, this often makes generated data resemble only a few modes of real data and rotate between modes. To alleviate these problems, we propose a novel training strategy to restrict greed in training of GAN. With help of our method, the generated samples can cover more instance modes with more stable training process. Evaluating our method on several representative datasets, we demonstrate superiority of improved training strategy on typical GAN models with different distance metrics.*

## 1. Introduction

Deep generative models can generate plausible data that resembles the real data, and this property makes it be widely researched recently. Variational auto-encoder (VAE) is an effective deep learning model to explicitly depict the generated data by maximizing a variational lower bound of the training data [1]. Generative adversarial networks (GAN), proposed by [2], offer a distinct and promising framework to implicitly synthesize images with adversarial mechanism. In general, generators in various GAN methods are trained to generate images from random noise. In this training process, generators can gradually obtain the ability to fool the adversarial-trained discriminator into predicting that generated instance are real. The property of GANs has been demonstrated in various computer vision and machine learning tasks, such as text-to-image transfer, super-resolution, and image domain transfer [3] [4] [5]. version.

Although high-quality results with meaningful coherence can be generated by variations of GAN, there are agreements that some problems have not been solved in training networks. During training, discriminator tries to assign higher data probability to true data and opposite lower probability to generated data. In fact, this can be thought of as a greedy strategy which attempting to minimize loss values of data batch in related distance metrics. As a consequence of this optimal goal, the samples produced by generator gradually tend to resemble only a few modes of real data, many other modes missed. This phenomenon is named as mode collapse [6], resulting in a lower diversity of generated samples compared with real data, which is shown in Figure 1. On the other hand, the training process of generator and discriminator often shows a state of oscillating instead of converging to a stable point. This is usually caused by the reason that discriminators are more powerful than generator, which provides useless instruction to generator.

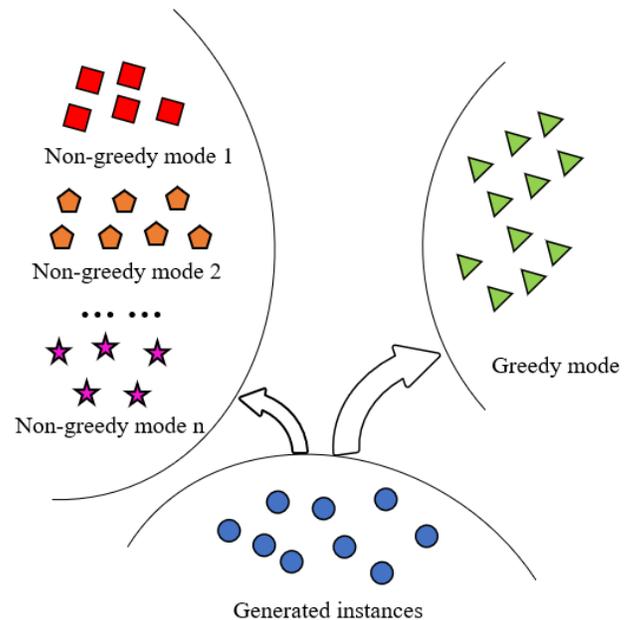

Figure 1: Illustration of missing modes problem. In traditional GAN models, the generated instances tend to resemble some modes (Greedy mode), which results in the missing of other data modes (Non-greedy mode 1, 2, …n).

The missing of generated modes is mainly caused by the unconstrained discriminator that only focuses on assigning high probability to real data manifold and low probability to generated instances manifold, without considering the biased inclination of the generator to push probability mass toward real data modes. Aiming for remedying the problems above, we propose a novel training strategy by restricting greed of GAN. Inspired by the $\varepsilon$ -greedy policies [23] which is widely used in reinforcement learning, our improved training strategy provides the generative network with the choices of not only exploitation (minimizing the loss function by hitting greedy data modes) but also exploration (induce the probability mass assigned by generator first to transition regions around various data modes).

Our method also achieves a promising trade-off be-tween diversity and quality of generative samples. According to the performance of GANs improved by our training strategy and traditional ones with different distance metrics, missing modes and instability are ameliorated with no harm to generative samples quality.

2. Related work

Owing to the efforts of researchers, GAN has achieved considerable improvement in recent years. Deep convolutional GAN is proposed to find a better convolutional structure, which reduces the instability and improves the quality of generated examples [7]. VAE is another widely used deep generative model. But VAE models often spread probability mass to places where it might not make sense, whereas GAN models may miss modes of the true distribution. Since these two streams of deep generative models possess advantages in different aspects, there are also some recent work trying to combine GAN and VAE in interesting ways. Energy-based GAN treats the discriminator from the prospective of energy function, using auto-encoder to replace discriminator [24]. The auto-encoder based discriminator makes the GAN get more stable con-vergence and more variable results. Additionally, GAN with denoising feature matching uses the denoising auto-encoder to match the higher-level features of the discriminator, achieving competitive performance [22]. Another variation named boundary equilibrium GAN (BEGAN) also applies the auto-encoder structure as a discriminator [8]. But it uses the reconstruction loss of auto-encoder as an optimization objective, driving reconstruction loss of generative samples close to that of real data. In BEGAN, proportional control theory is adopted to maintain the equilibrium and avoid an instability training by much more emphasize on generator during gradient descent.

Being similar with its success on unsupervised learning, GAN is extended to the field of semi-supervised learning with labels. Conditional GAN conditions generator and discriminator on some extra information such as class labels [9]. By feeding class labels into both the generator and discriminator during training, it could generate samples conditioned on the class label. Mutual information between latent variables and observation is exploited in Info-GAN [10]. By maximizing the mutual information, Info-GAN is encouraged to learn interpretable and meaningful representations, which successfully disentangles various features of training data.

To improve GAN in more theoretical ways, different probability metrics have received much attention from researchers. In addition to the widely used Kullback-Leibler (KL) divergence, many other metrics have also been researched recently. [11] points out that the generative-adversarial approach is a special case of an existing more general variational divergence estimation approach. And it derives the GAN training objectives for all f-divergences, including KL and Pearson divergences. To better reflect the underlying geometry between outcomes, Wasserstein metric is proposed to indicate the distance between two distributions by [12] [21]. [13] proposes a novel GAN based on the Cramér distance that own the properties of unbiased sample gradients and sum invariance and scale sensitivity. The Cramér distance combines the advantages of the Wasserstein and KL divergences and achieve a more satisfying performance in their experiments.

In addition, GAN is gradually applied into translating multimodal information. Based on GAN structure, [3] develops a novel deep architecture that conditions on text description, translating visual concepts from characters to images. In experiments, the model indeed learns to generate reasonable images given detailed text description. DiscoGAN, proposed in [14], learns to discover relations be-tween different domains, and successfully implements style transfers from one domain to another with keeping the main characteristic unchanged. Many general problems often exist in discrete outputs of generative models, such as the difficulty of gradient propagation. To solve the problems, [15] proposes a sequence generation framework that performs gradient policy update with Monte Carlo search to pass the gradient back.

However, both theoretical improvements and emerging applications of GANs above follow the same general guideline in their training procedures: minimizing the discrimination loss greedily under related metrics (f-divergences, including KL, Pearson divergences or earth mover distance), guiding the generator to output plausible instances. Even though state-of-the-art Wasserstein metric based method can alleviate the problem of unstable training, it still misses some data modes in generated examples. This can demonstrate the "greedy" training strategy can-not cover the problem of missing modes.

3. Our method

In this section, we first review the original GAN and reveal the problems often existing during training process. Then we analyze the causes behind them and give an explanation about how our improved training strategy alleviates the greed of traditional GANs (the typical DCGAN and WGAN).

### 3.1. Generative adversarial network

GAN is an adversarial framework, containing two neural networks, discriminator $D(\cdot\ ;\ \theta_D): \mathcal{X} \to [0, 1]$ and generator $G(\cdot\ ;\ \theta_G)\ \mathcal{Z} \to \mathcal{X}$, where $\theta_D$ and $\theta_G$ are the trainable parameters of discriminator and generator respectively. The generator transforms a random vector $z$ in the latent space to a sample in the data space. The discriminator maps a sample $x$ to an estimated probability that $x$ comes from the real data. The standard GAN loss function is optimized as:

$$\min_{\theta_G} \max_{\theta_D} \mathbb{E}_{x \sim p_{data}}\left[\log\left(D(x;\theta_D)\right)\right] + \mathbb{E}_{z \sim p_z}\left[\log\left(1 - D\left(G(z;\theta_G);\theta_D\right)\right)\right] \quad (1)$$

where $P_{data}$ is the real data distribution, $x$ is the data variable, $z$ is randomly generated in latent space $\mathcal{Z}$. Intuitively, the purpose of discriminator is to distinguish generated samples from the real samples, while generator tries to generate plausible samples to fool the discriminator. All along the training, $\theta_D$ and $\theta_G$ are updated alternatively with an adversarial status between two networks. For generator, the updated gradient is back propagated from the discriminator. During optimizing, both generator $G$ and discriminator $D$ improve their abilities and gradually converge to an expected point where $G$ is able to generate plausible samples resembling real data, and $D$ could figure out rather fake samples.

### 3.2. Motivation

Here we give a more radically analysis toward the missing mode phenomenon and training instability. The optimization process can be considered as that generator pushes generation manifold close to real data manifold that has a high probability assigned by discriminator greedily. In theory, if the generative samples and real data are in a low dimensional space and the loss function Eq. (1) is convex in $\theta_G$ and concave in $\theta_D$, the probability mass will be distributed to various modes in a sparse way. However, the unsatisfaction of conditions above makes the networks not so perfect as in theory. Assuming discriminator assigns a high value both to all different modes, imperfect generator still has the inclination to push generative samples toward some modes, leaving other modes missed, because the missing modes are not as attractive as non-missing ones considering the purpose of generator is to maximize the $D(G(z; \theta_G); \theta_D)$ rather than cover all the data modes.

Another problem is the instability during training. In high dimensional space, the two low dimensional manifolds (data manifold and generation manifold) are more likely to be disjoint [12]. In standard GAN models, as the proceeding of training, optimal discriminator will assign 1 to data manifold and 0 to generation manifold accurately which results the disjoint of them. In that cases, generator will not get useful gradient from discriminator to update $\theta_G$, which causes the oscillating and instability. Our solution to the problems above is detailed in the next section.

### 3.3. Restricting greed in training of GAN

In reinforcement learning, $\varepsilon$-greedy policies mean that methods choose an action that has maximal estimated action value with the probability of $1-\varepsilon$ or an explored action with the probability of $\varepsilon$. When it comes to the training of GAN, the greedy action is equal to minimizing loss function on data batch in one iteration.

To provide more opportunities for exploration, we introduce a relaxation $R$ to discriminator loss in GAN. The related discriminator loss and generator loss are shown as the equations below:

$$L_D = \mathbb{E}_{x \sim p_{data}}\left[\log\left(D(x;\theta_D)\right)\right] + \mathbb{E}_{z \sim p_z}\left[\log\left(1 - D\left(G(z;\theta_G);\theta_D\right)\right)\right] + R \quad (2)$$

$$L_G = \mathbb{E}_{z \sim p_z}\left[\log\left(D\left(G(z;\theta_G);\theta_D\right)\right)\right] \quad (3)$$

$$R = \lambda \mathbb{E}_{\hat{x} \sim p_{\hat{x}}}\left[\log\left(D(\hat{x};\theta_D)\right)\right] \quad (4)$$

$$\hat{x} = (1-t)x + ty \quad (5)$$

Eq. (2) represents the discriminator loss $L_D$, Eq. (3) stands for the generator loss $L_G$. And they are optimized in the training process to obtain the parameters of discriminator network and generator network. Eq. (4) is the relaxation with Eq. (5) where $t \in [0, 0.5]$, $x$ is sampled from $P_{data}$, $y$ is sampled from $P_G$ (distribution of generated data), $\lambda$ is a coefficient standing for the degree of exploration. And $\hat{x}$ defined as a point in a straight line between points sampled from data distribution $P_{data}$ and generated data distribution $P_G$. The range of $t$ is determined from our empirical experiments with more generalization to be [0, 0.5]. Since $t \in [0, 0.5]$, can be restricted to be closer to $P_{data}$. In standard GAN, the optimal discriminator is in the form of Eq. (6) [12]:

$$D^*(x) = \frac{P_{data}(x)}{P_{data}(x) + P_{G(z)}(x)} \quad (6)$$

When $x$ is in the area between generated instances and real data mode with $P_{data}(x) \to 0$, $D^*(x) \to 0$, discriminator cannot pass sufficient information to generator which makes some data modes unaccessible.

In our training strategy, we have the optimal discriminator in the form of Eq. (7):

$$D^*(x) = \frac{P_{data}(x) + \lambda P_{\hat{x}}(x)}{P_{data}(x) + P_{G(z)}(x) + \lambda P_{\hat{x}}(x)} \quad (7)$$

when $x$ is in the area of $\hat{x}$, even though $P_{data}(x) \to 0$, $D^*(x)$ can provide instructive information for generator as Eq. (8):

$$D^*(x) = \frac{\lambda P_{\hat{x}}(x)}{P_{G(z)}(x) + \lambda P_{\hat{x}}(x)} \quad (8)$$

Our relaxation encourages the discriminator to additionally assign intermediate values $D(\hat{x}; \theta_D)$ lower than $D(x; \theta_D)$ but higher than $D(G(z; \theta_G); \theta_D)$, to the distribution $P_{\hat{x}}$. In the other words, the transition regions between the real data modes and generated samples also present estimated probability given by $D$ ranging in ($D(G(z; \theta_G); \theta_D), D(x; \theta_D)$) as is illustrated in Figure 2.

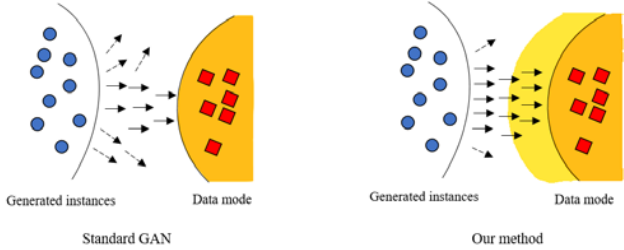

Figure 2: Illustration of our method. In standard GAN, discriminator only assigns higher values (orange region) to $P_{data}$(data mode in red square). The optimization of our relaxation regularizer sets a transition region $P_{\hat{x}}$ (yellow region) with intermediate values. Thus, the generated instances can approach the missing data mode more easily.

And thus, our method indeed brings a broader feasible region when generator tries to maximize $D(G(z; \theta_G); \theta_D)$. Intuitively, for missing modes, it is more accessible to push possibility mass to the transition region with intermediate estimated possibility than directly push possibility mass to modes, because in former process, generator could improve $D(G(z; \theta_G); \theta_D)$ more easily. So, the generative samples are induced to transition region in each iteration, and transition region also gets closer to real data. And thus, more diversity of generated samples can be achieved and mode collapse is alleviated. Additionally, during the optimization, decay coefficient of our relaxation is decayed gradually, which means our training strategy gradually turns to exploitation after enough exploration, resulting to convergence without sacrificing the quality of generated samples. The phenomenon can be demonstrated in the experiments below.

Another benefit of our method is that during training of discriminator, the frontier of real data manifold is "extended" by assigning the region around data manifold with intermediate values. So, generation manifold is less likely to be disjoint with the "extended" data manifold, which increases the stability of training.

## 4. Experiments

In this section, the effectiveness of our method on DCGAN and WGAN in alleviating the mode collapse and instability training is demonstrated by experiments on several datasets. When it comes to WGAN, our strategy is applied with relaxation in the form of Eq. (9):

$$R = \lambda \mathbb{E}_{\hat{x} \sim p_{\hat{x}}}\left[D(\hat{x}; \theta_D)\right] \quad (9)$$

### 4.1. Evaluation methods in experiments

Evaluation in generative models is particularly challenging and various methods have been proposed. Independent Wasserstein critic [16], inception score [17] and multi-scale structural similarity [18] are adopted in our experiments.

Independent Wasserstein critic is a robust critic to evaluate both the quality and diversity of samples. It first trains a Wasserstein GAN and then extracts its discriminator as an independent critic $f$. During training in other experiments, the approximate distance between generated samples $G(z; \theta_G)$ and $x$ can be measured by:

$$W = f(x) - f(G(z; \theta_G)) \quad (10)$$

where lower values of $W$ mean more diversity and higher quality of generative samples.

Inception score is widely used as an assessment of quality of images. It computes the average KL divergences between softmax output of a trained classifier of the labels and marginal distribution over all samples. So high inception scores are presented if the generated samples that are sent to classifier (usually an Inception network trained on the ImageNet dataset) have high quality. However, for our task of measuring missing modes, inception score is not so powerful because it only focuses on the diversity with regards to labels and quality.

Multi-scale structural similarity (MS-SSIM) is an effective image similarity metric, ranging between 0 and 1. Given two groups of images, the MS-SSIM could figure out how similar they are. Thus, it is often used in measuring the diversity by presenting a lower value for more varied images.

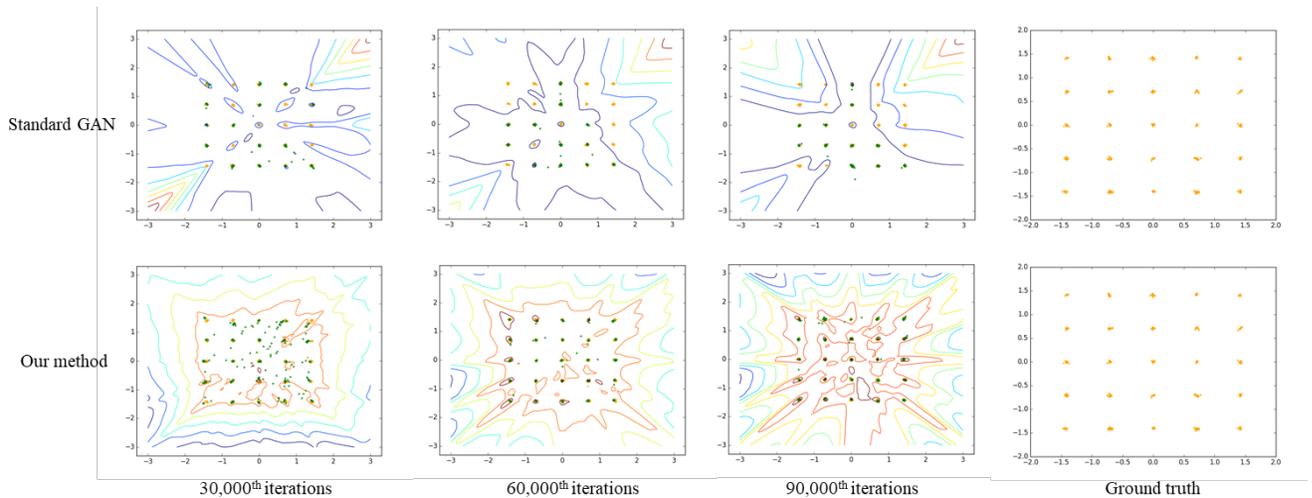

Figure 3: Performance of standard GAN and our model in 2-D 25 Gaussians. The rightmost column presents the real data distribution (orange dots). The left three columns show the generation distribution (green dots) changing with training iterations. The contour lines indicate probability output of discriminator in 2-D space. The top row stands for vanilla GAN, where the green dots cannot cover all the orange dots, varying in the modes of covered dots. While the bottom row for our method shows a rather stable training and cover of all modes. Since the 2-D Gaussians data is too simple for WGAN, both traditional and our training strategy can achieve the competitive performance. Owing to the limitation of length, we do not list them in this figure.

### 4.2. Mixture of Gaussians dataset

The dataset is sampled from mixture of 25 Gaussians arranged in a rectangle. The structure we use is simple: both generator and discriminator have three ReLU layers and one fully-connected layer. Both standard GAN and our method are optimized by Adam optimizer. Figure 3 shows the performance of them. The samples generated by standard GAN just could cover several modes and often rotate among different modes. With our relaxation regularizer, generated samples are spread to all modes and gradually reach a rather stable status.

### 4.3. CelebA Dataset

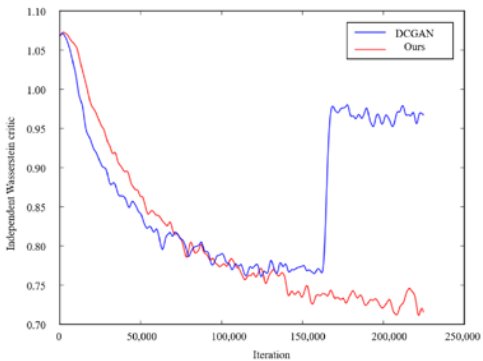

Figure 4: Illustration of Wasserstein critic during training in our experiments.

CelebA is a large-scale face dataset with more than 200K images [19]. In experiments, DCGAN and WGAN are trained by both traditional strategy and our proposed one to be applied to generate 64 × 64 center-cropped CelebA images. Our method uses DCGAN architecture as structure baseline, with default optimizer hyper-parameters recommended in [7] and an exponential decay of coefficient $\lambda$ in this task.

We use independent Wasserstein critic and multi-scale structural similarity to evaluate the diversity of generated samples and stability in training.

As is shown in Figure 4, the DCGAN often suffers severe mode collapse after training too many iterations (explosion of independent Wasserstein critic) while our method could alleviate this problem and get lower values which means more stable and no harm to quality. Since WGAN has been demonstrated to be more stable in previous work, we do not analyze the stability of it. Following the subjective evaluation in standard GAN models, we list generated images of both standard GANs and our method in Figure 5. Comparing the generated instances of these two methods, we can find that our method can maintain the quality of generated images. Meanwhile, our training strategy can assist the GANs to generate more data modes.

We also list the MS-SSIM performance of our experiments in Table 1. As is shown in this table, our method also outperforms traditional DCGAN and WGAN, obtaining MS-SSIM values which are fairly close to real data. It implies our method certainly scatters the generated data to more modes and increases the diversity of them.

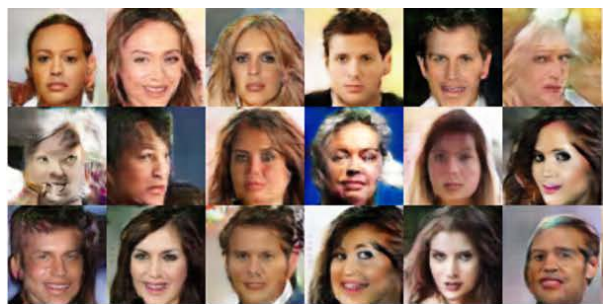
DCGAN

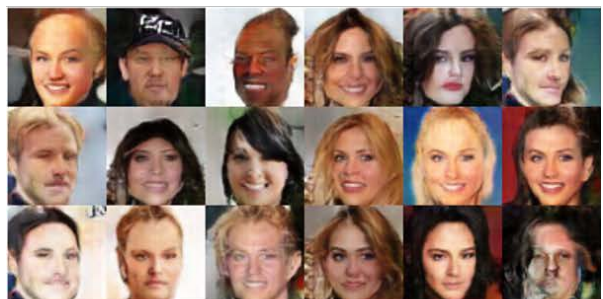
DCGAN-Ours

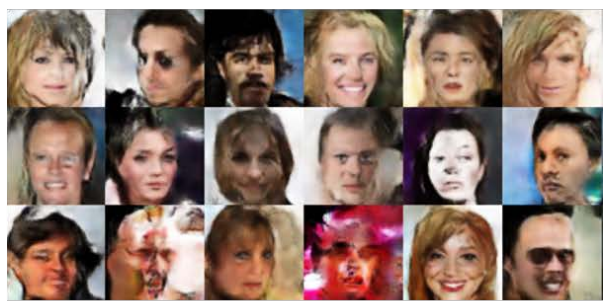
WGAN

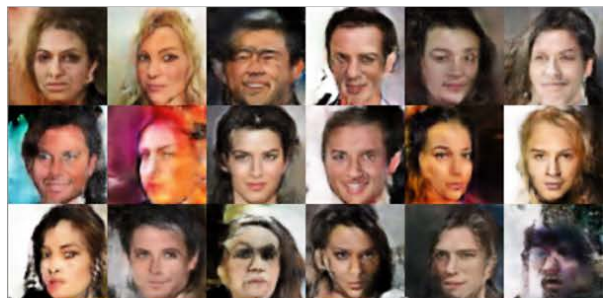
WGAN-Ours

Figure 5: Generated instances of DCGAN and WGAN which are trained by traditional strategy and ours.

Table 1: MS-SSIM performance on CelebA dataset

| *Method* | *MS-SSIM* |
|---|---|
| DCGAN | 0.3150 |
| DCGAN-Ours | **0.3083** |
| WGAN | 0.2945 |
| WGAN-Ours | **0.2913** |
| Ground truth | 0.2940 |

4.4. CIFAR-10 Dataset

CIFAR-10 is a dataset of 32 × 32 natural images [20], and it is widely studied in many researches. Our model also has DCGAN baseline with recommended hyper-parameters and exponentially decayed coefficient $\lambda$.

The images in CIFAR-10 are extremely variable. So, for better measuring the diversity of generated images, we preform experiments on images of every class in CIFAR-10 respectively, and thus MS-SSIM could be more efficient to indicate intra-class images diversity. The performance of models regarding to quality and diversity is shown in Table 2 and Table 3.

Table 2: Comparison of MS-SSIM in traditional DCGAN and DCGAN-Ours.

| **Category** | *DCGAN* | *DCGAN-Ours* |
|---|---|---|
| Airplane | 0.1116 | **0.1001** |
| Automobile | 0.1169 | **0.1125** |
| Bird | 0.0789 | **0.0709** |
| Cat | 0.0607 | **0.0589** |
| Deer | 0.0854 | **0.0808** |
| Dog | 0.0572 | **0.0542** |
| Frog | 0.0993 | **0.0990** |
| Horse | 0.0712 | **0.0696** |
| Ship | 0.1416 | **0.1299** |
| Truck | **0.1478** | 0.1521 |

Table 3: Comparison of MS-SSIM in traditional WGAN and WGAN-Ours.

| Category | WCGAN | WGAN-Ours |
|---|---|---|
| Airplane | **0.0899** | 0.0936 |
| Automobile | 0.1038 | **0.1019** |
| Bird | 0.0647 | **0.0635** |
| Cat | 0.0537 | **0.0446** |
| Deer | 0.0610 | **0.0603** |
| Dog | 0.0474 | **0.0457** |
| Frog | 0.0498 | **0.0452** |
| Horse | 0.0636 | **0.0556** |
| Ship | 0.0691 | **0.0681** |
| Truck | **0.1395** | 0.1438 |

From above Table 2 and Table 3, our proposed method gets lower values in 9 classes among all 10 classes for DCGAN and in 8 classes among all 10 classes for WGAN, meaning more diversity and less missing modes of generated samples. Thus, our method indeed shows impressive improvement toward traditional GAN models.

Table 4: Comparison of inception scores in experiments.

| Category | DC-GAN | DCGAN-Ours | W-GAN | W-GAN-ours |
|---|---|---|---|---|
| Airplane | **4.46** | 4.45 | 4.54 | **4.56** |
| Automobile | 3.72 | **3.74** | **3.47** | 3.35 |
| Bird | **4.44** | 4.32 | **4.77** | 4.72 |
| Cat | **4.21** | 4.15 | 4.41 | **4.51** |
| Deer | 3.98 | **4.11** | 4.16 | **4.26** |
| Dog | 4.79 | **4.89** | 4.97 | **5.04** |
| Frog | **3.63** | 3.46 | **3.63** | 3.59 |
| Horse | **4.43** | 4.30 | **4.39** | 4.37 |
| Ship | **3.68** | 3.58 | 3.82 | **3.87** |
| Truck | 3.19 | **3.20** | 3.07 | **3.10** |

As for the quality of images, inception score is used to assess our experiments. Shown in Table 4, our method achieves superior performance on four classes, which means approximate results compared with traditional DCGAN and WGAN. And it indicates that our method does not harm to generated samples quality. Some of the generated instances are listed in Figure 6.

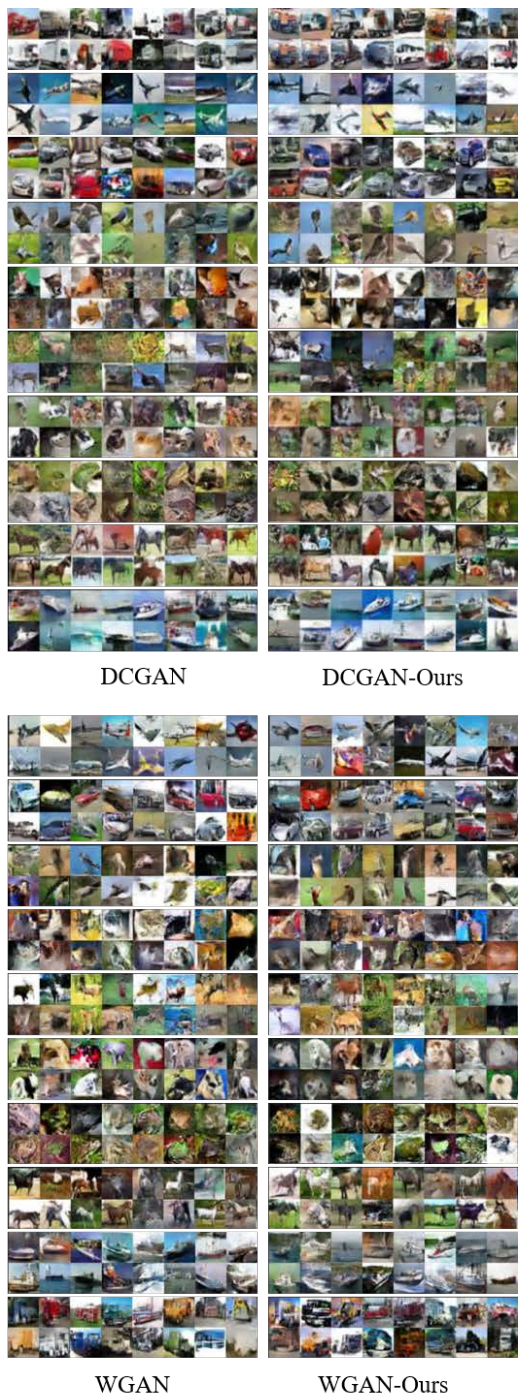

DCGAN      DCGAN-Ours

WGAN      WGAN-Ours

Figure 6. Generated samples on CIFAR-10 dataset.

## 5. Conclusion

Deep learning method is the research hotspot of computer vision and machine learning areas. It has achieved superior performance on a variety of tasks [25] [26] [27] [28] [29] [30] [31]. Deep generative model is one of the core issues in deep learning research. Although GAN can generate decent samples in many tasks, the problems of mode collapse and instability training are remained deficiencies that need to be solved urgently. In our work, an efficient method is proposed to alleviate missing mode problem and stabilize training. Specifically, an exploration strategy is introduced to restrict the greed of GAN training, making the missing modes more accessible for generative samples. In our experiments, the evaluation metrics for diversity and quality are used to assess the performance of models. With the help of our improved training strategy, both typical DCGAN and WGAN outperform their standard versions with more modes covered and more stability in training. For future work, the generalization of our method is planned to be explored by applying it to more variations of GANs.